\documentclass[preprint,12pt,authoryear]{elsarticle}
\usepackage{lineno,hyperref}
\usepackage{amsmath}
\usepackage{algorithm}
\usepackage{algorithmic}
\usepackage{array}
\usepackage{multirow}
\usepackage{booktabs}
\usepackage{graphicx}
\usepackage{amssymb}
\usepackage{subfigure}

\usepackage{stfloats}
\usepackage{url}
\usepackage{times}

\usepackage{caption}

\usepackage{booktabs}
\usepackage{array, caption, threeparttable}
\usepackage[font=small,labelfont=bf,labelsep=none]{caption}
\usepackage{amssymb}

\journal{Computers in Industry}

\begin{document}

\captionsetup[figure]{labelfont={bf},labelformat={default},labelsep=period,name={Fig.}}

\begin{frontmatter}

\title{A Lightweight and Accurate Recognition Framework for Signs of X-ray Weld Images}
\author{Moyun Liu\fnref{a}}
\ead{lmomoy@hust.edu.cn}

\author{Jingming Xie\corref{cor1}\fnref{a}}
\ead{xjmhust@hust.edu.cn}

\author{Jing Hao\fnref{a}}
\ead{jinghao@hust.edu.cn}

\author{Yang Zhang\fnref{b}}
\ead{yzhangcst@gmail.com}

\author{Xuzhan Chen\fnref{a}}
\ead{chenxuzhan@hust.edu.cn}

\author{Youping Chen\fnref{a}}
\ead{ypchen@hust.edu.cn}

\cortext[cor1]{Corresponding author}

\address[a]{School of Mechanical Science and Engineering, Huazhong University of Science and Technology, Wuhan, 430074, China}

\address[b]{School of Mechanical Engineering, Hubei University of Technology, Wuhan, 430068, China}

\begin{abstract}
X-ray images are commonly used to ensure the security of devices in quality inspection industry. The recognition of signs printed on X-ray weld images plays an essential role in digital traceability system of manufacturing industry. However, the scales of objects vary different greatly in weld images, and it hinders us to achieve satisfactory recognition. In this paper, we propose a signs recognition framework based on convolutional neural networks (CNNs) for weld images. The proposed framework firstly contains a shallow classification network for correcting the pose of images. Moreover, we present a novel spatial and channel enhancement (SCE) module to address the above scale problem. This module can integrate multi-scale features and adaptively assign weights for each feature source. Based on SCE module, a narrow network is designed for final weld information recognition. To enhance the practicability of our framework, we carefully design the architecture of framework with a few parameters and computations. Experimental results show that our framework achieves 99.7\% accuracy with 1.1 giga floating-point of operations (GFLOPs) on classification stage, and 90.0 mean average precision (mAP) with 176.1 frames per second (FPS) on recognition stage.
\end{abstract}

\begin{keyword}
Convolutional neural network \sep signs recognition \sep X-ray weld images, lightweight.
\end{keyword}

\end{frontmatter}


\section{Introduction}
Manufacturing industry relies on X-ray images to monitor weld quality in daily production, because they have ability to reflect the internal condition of artifacts~\citep{malarvel2017anisotropic}. Some signs printed on X-ray images include cross mark and weld information such as the date of the photograph, the serial number of the artifact and the mark of image quality indicator (IQI). Cross mark is used to show the pose of images, and weld information needs to be stored into digital system for tracing images. Therefore, an automatic signs recognition framework is vital for an advanced digital X-ray weld image system. These signs are produced by some moveable types whose material is plumbum, and they are selectively placed on the top of weld metal manually. Finally, these signs would be projected into image through X-ray detection apparatus.

\begin{table}[!t]
\renewcommand{\arraystretch}{1.1}
\centering
\caption{List of abbreviations}
\begin{tabular}{cccccc}
\toprule
Abbreviation & Full name \\
\hline
BDB          & basicblock-downsample-basicblock \\
\hline
BN           & batch normalization \\
\hline
CBR          & convolutional-batch normalization-rectified linear unit   \\
\hline
CNNs         & convolutional neural networks  \\
\hline
Conv         & convolution  \\
\hline
FN           & false negative  \\
\hline
FP           & false positive \\
\hline
FPS          & frames per second  \\
\hline
GFLOPs       & giga floating-point of operations \\
\hline
GRNet        & group convolution-based resnet  \\
\hline
GYNet        & greater-YOLO network   \\
\hline
IQI          & image quality indicator  \\
\hline
Madds        & multiply-add operations \\
\hline
mAP          & mean average precision   \\
\hline
Params       & parameters   \\
\hline
ReLU         & rectified linear unit   \\
\hline
SCE          & spatial and channel enhancement \\
\hline
SGD          & stochastic gradient descent  \\
\hline
TP           & true positive   \\
\bottomrule
\end{tabular}
\label{tab:abbreviations}
\end{table}

\begin{figure}[!t]
  \centering
  \includegraphics[width=5in]{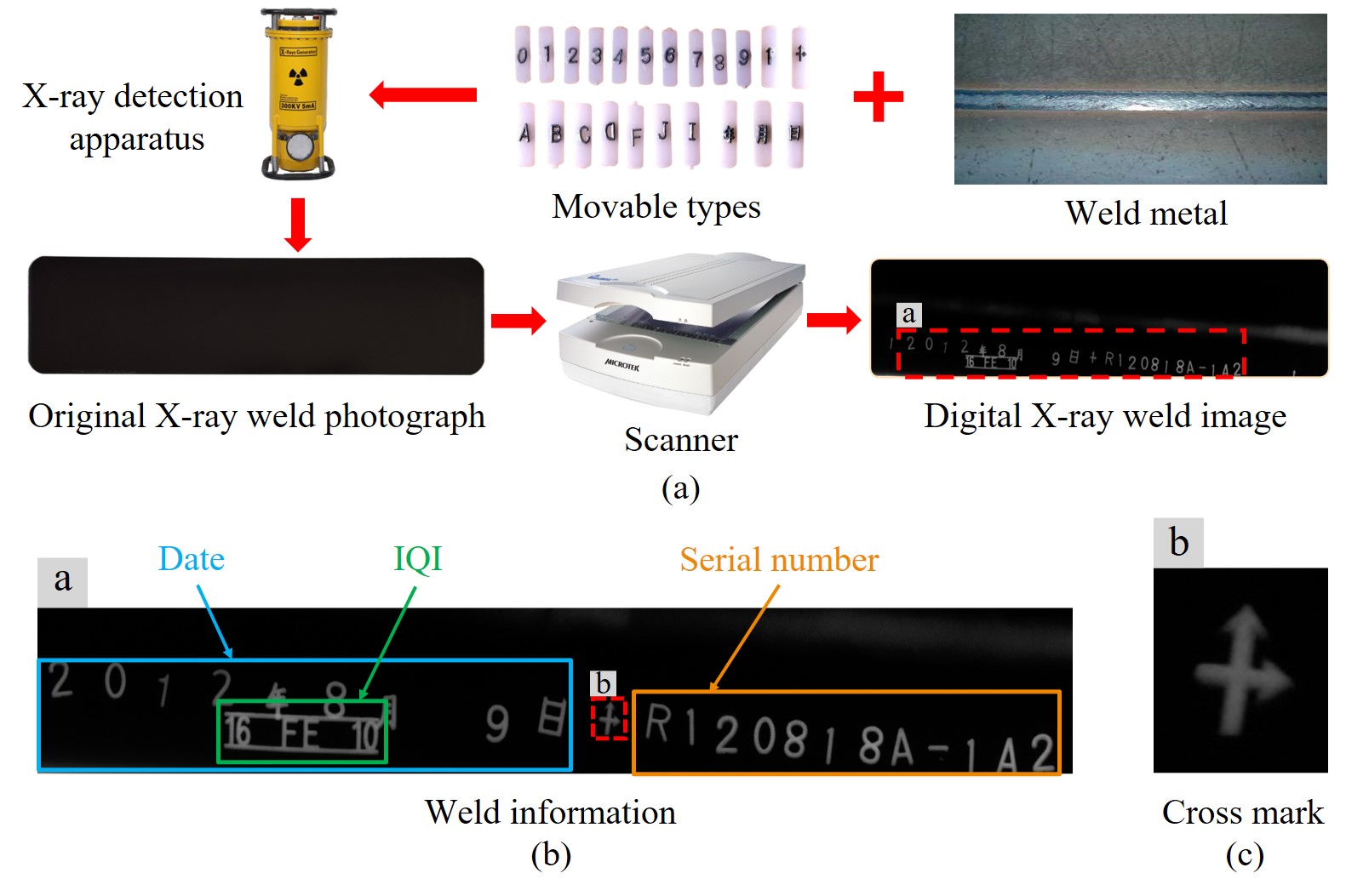}
  \caption{Digitization of X-ray weld image. (a) Movable types are placed on the top of weld metal, and original X-ray weld photograph is produced by X-ray detection apparatus. Finally, the photograph is transformed to digital image by scanner. (b) Weld information mainly contains the date of the photograph, the serial number and IQI. (c) Cross mark can reflect the pose of weld image.}
  \label{fig:Digital}
\end{figure}

To better save and observe, original weld photographs are scanned to digital images as shown in Fig.~\ref{fig:Digital}. There is a cross mark on each image, and it is usually unordered. Only the mark showing right\&up means the correct direction, and it is necessary to redirect the image based on the classification result of cross mark. Weld information printed on image is required to be further recognized after completing the forementioned classification task. The categories of these information are mainly numbers, letters and some marks. We regard the information recognition as object detection task, which has been studied for many years in deep learning field.

In recent years, deep learning approaches are facing vigorous development, among which convolutional neural networks (CNNs)-based methods have made excellent achievements in various image tasks. To exploit the potential of CNNs, researchers have proposed many superior network structures for image classification~\citep{simonyan2014very, he2016deep} and object detection~\citep{ren2015faster, redmon2016you}. However, to our knowledge, these methods are more used for defects on weld images~\citep{yaping2019research, duan2019automatic, dong2021automatic}, but not for signs recognition. Besides signs, the foreground contents also include weld region and noises, and they have a variety of scales. Hence, achieving recognition on X-ray weld images is a challenging task because the context of image is complex. In general, multi-scale features fusion~\citep{bell2016inside, lin2017feature} is introduced to address the scale diversity problem. This strategy allows the extracted feature maps to obtain features at different scales simultaneously, and it is also helpful to predict objects of different sizes~\citep{redmon2018yolov3}. However, since size distribution of weld information is consistent and single, the features at one scale are often more important than at other scales, so it is crucial to assign weights for different feature sources at weld information recognition task. There are many existing methods designed for enhancing the multi-scale representation ability of network~\citep{he2015spatial, chen2017deeplab}. However they usually simply add or concatenate feature maps from different scale sources, without ranking their importance and assigning them with different weights.


In this paper, we propose a signs recognition framework for X-ray images to accomplish the above tasks. Our framework is compact and high-performant, consisting of two CNNs, i.e., Group convolution-based ResNet (GRNet) for cross mark classification and Greater-YOLO network (GYNet) for weld information recognition. Based on the residual block of ResNet, we design a shallow backbone for GRNet, and group convolution~\citep{howard2017mobilenets} is introduced to reduce the parameters and computations. Inspired by the efficient structure of Tiny-YOLO v3~\citep{redmon2018yolov3}, we propose a more narrow GYNet based on a novel spatial and channel enhancement (SCE) module. SCE module firstly integrates features from multiple scales, and then adaptively weights them according to their contributions. To validate the effectiveness of our framework, we conduct extensive experiments on our datasets. Experimental results show that our framework achieves high performance with fast speed and a few parameters, compared with the state-of-the-art methods.


In summary, this work makes the following contributions.
\begin{itemize}
  \item We design a compressed and accurate framework to fulfill the signs recognition of weld images with fast speed and high performance.
  \item A elaborate backbone for GRNet is proposed, and it is designed with a few layers based on group convolution.
  \item We propose a narrow and light GYNet, in which a novel SCE module is introduced to complement feature information at different scales and weight them adaptively.
  \item The experimental results show that our methods achieve fast speed and accurate prediction compared with state-of-the-art models.
\end{itemize}

The rest of this paper is organized as follows. Section 2 introduces some related works about CNNs of classification, detection and multi-scale features fusion methods. Section 3 presents our framework in detail. All experimental results are shown and discussed in Section 4. Finally, we conclude this paper in Section 5. A list of abbreviations is listed in Table~\ref{tab:abbreviations}.

\section{Related works}
To the best of our knowledge, there is no research on signs recognition of weld images in the past. We regard it as the task of image classification and detection, which has been studied for many years, and many excellent works have been proposed.

\subsection{Image Classification and Object Detection}

The use of convolutional neural network for image classification can be traced back to the 1990s, when the proposed LeNet-5~\citep{lecun1998gradient} laid the foundation of CNNs. AlexNet~\citep{krizhevsky2012imagenet} won the first prize ImageNet competition in 2012, and triggered the research of CNNs. VGG~\citep{simonyan2014very} reduces the size of the filter to $3\times3$ and deepens the network depth, greatly improving the classification accuracy on the ImageNet dataset. ResNet~\citep{he2016deep} increases the potential of CNNs by introducing residual connection, which solves the problem of gradient disappearance in the training process and makes it possible to design deeper networks.

As for object detection, CNNs can be divided into one-stage and two-stage detector. The biggest difference between them is that the latter generates regional proposal, while the former does not. The classical two-stage detectors, such as Faster R-CNN~\citep{ren2015faster}, Cascade R-CNN~\citep{cai2018cascade} and Libra R-CNN~\citep{pang2019libra} firstly generate a set of region proposals, and they will be classified and regressed bounding box at the end. Two-stage methods usually can achieve more accurate prediction results, but they also need more computation resources, and are not satisfactory on detection speed. One-stage models are more applicable when tasks have requirements on inference speed. YOLO~\citep{redmon2016you, redmon2017yolo9000, redmon2018yolov3}, SSD~\citep{liu2016ssd}, RetinaNet~\citep{lin2017focal} are typical one-stage CNNs. Although they have higher efficiency, the lack of region proposal step makes them not accurate enough compared with two-stage network in most cases.

In addition, many lightweight classification and detection networks are designed to enhance the practicability of CNNs. A novel Fire Module was proposed in SqueezeNet~\citep{iandola2016squeezenet}. This module reduces parameters by using $1\times1$ convolution to replace $3\times3$ convolution. The MobileNet~\citep{2018MobileNetV2} series networks proposed depthwise separable convolution that can reduce the model complexity. ShuffleNet~\citep{zhang2018shufflenet, 2018ShuffleNet} changes the channel order of feature maps, which enables cross-group information flow. A cheap operation is introduced in GhostNet~\citep{han2020ghostnet}, which has fewer parameters while obtaining the same number of feature maps.

\subsection{Multi-scale Features Fusion}
Feature fusion is a very effective strategy to achieve feature complementarity among different layers of CNNs. The original fusion way is simply adding or concatenating multi-scale features~\citep{bell2016inside}, which achieves improvement on performance to some extend. To obtain better approaches, more fusion strategies are exploited. SSD~\citep{liu2016ssd} and MS-CNN~\citep{cai2016unified} directly combine prediction results of the feature hierarchy. Feature Pyramid Networks~\citep{lin2017feature} constructs a top-down structure to fuse feature layers of different scales, and produces enhanced feature maps for final classification and localization prediction. Recently, some researches have found that multi-scale fusion based on different receptive fields can greatly improve the performance of CNNs. For example, SPP~\citep{he2015spatial} generates multiple groups of feature maps through pooling operations with different kernel size. Similarly, ASPP~\citep{chen2017deeplab} achieves above goal by using atrous convolutions with different dilation rates. In spite of success, the current fusion methods do not consider which scale is more important for the final prediction. The essence of these strategies is to treat all scales equally.


An incredible recognition framework requires an outstanding baseline, and advanced feature fusion method. To achieve this goal, we design our classification network based on residual structure~\citep{he2016deep} and the convolution method used in MobileNet~\citep{2018MobileNetV2}. Inspired by the fast speed of one-stage model, we propose our recognition network based on Tiny-YOLO v3~\citep{redmon2018yolov3}. Moreover, a new feature map fusion method named SCE is proposed, and it is used to improve the multi-scale representation ability of recognition network.

\section{Method}
\begin{figure*}[!t]
  \centering
  \includegraphics[width=5in]{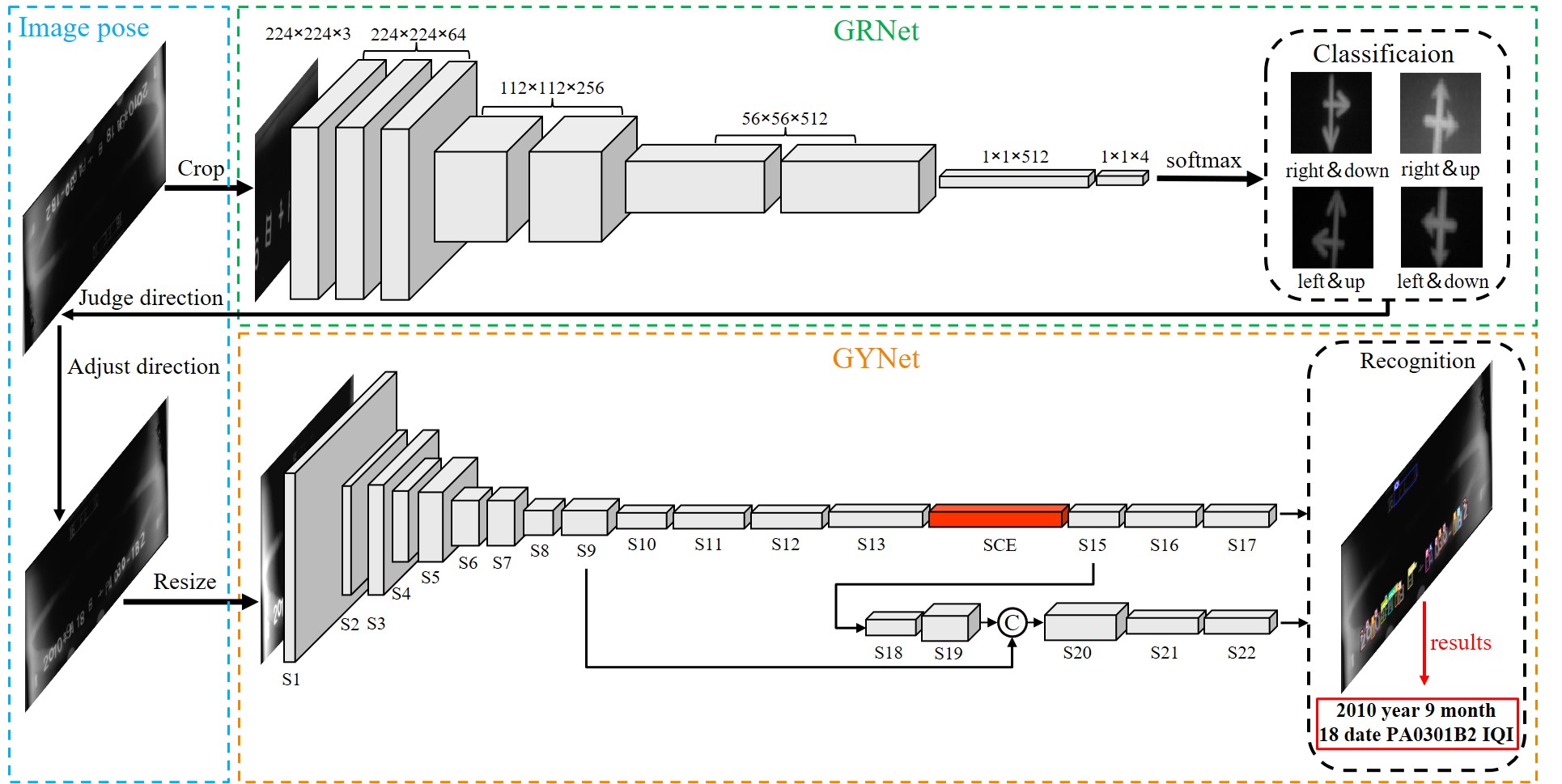}
  \caption{The architecture of weld information recognition framework. Our framework contains two networks, GRNet for cross mark classification and GYNet for weld information recognition. The image is first input into GRNet, and the category of the cross mark is obtained through 14 layers of convolution. Based on this classification result, the image is adjusted to correct pose. And then, image is extracted feature by lightweight GYNet which owns two prediction heads, to output the information content printed on X-ray weld image.}
  \label{fig:the structure of framework}
\end{figure*}

The proposed framework consists of two CNNs, i.e., GRNet for cross mark classification and GYNet for weld information recognition. The architecture of our framework is represented in Fig.~\ref{fig:the structure of framework}. GRNet is a lightweight yet effective classifier with only 14 convolution (Conv) layers. GYNet is a compressed but high-performing network designed by a few number of channels on high-level layers. In this section, we will explain the detailed structures of GRNet and GYNet.

\subsection{GRNet for Cross Mark Classification}
The final purpose of our framework is to recognize the information of X-ray weld images. However, the pose of these digital images is random and casual in actual production. Thus we need to classify the direction mark, i,e, the cross mark at first, and then adjust image to correct pose. A compact and efficient classification network can redirect faster, improving the overall efficiency of recognition framework. The obvious insight is to build the classifier with a few layers and lightweight convolution way. ResNet~\citep{he2016deep} is a successful series of classification CNN, but there is still room for further optimization in terms of the number of layers and internal modules. To achieve this goal, we propose a novel GRNet.

The backbone of our GRNet has 9 modules with only 14 Conv layers, and its architecture is shown in Table~\ref{GRNet}. The input images with $224\times224\times3$ resolution are fed into a Conv-BN-ReLU (CBR) module. CBR contains Conv layer with the kernel of $3\times3$ followed by Batch Normalization (BN)~\citep{ioffe2015batch} and Rectified Linear Unit (ReLU)~\citep{nair2010rectified}. And then GRNet employs a BasicBlock which is a residual module including a CBR module and $3\times3$ Conv followed by BN layer. The input and output feature maps of BasicBlock will be fused by element-wise adding. BasicBlock-Downsample-BasicBlock (BDB) is used to downsample the feature maps, and it has two branches. One of them is a BasicBlock whose stride of the Conv layer in CBR module is 2. Another branch employs a MaxPool layer whose size is $2\times2$ and stride is 2, followed by a $3\times3$ Conv with stride of 1 and BN layer. Finally, the obtained feature maps by two branches are added element-wise as well.

\begin{table}[!t]
\renewcommand{\arraystretch}{1.1}
\centering
\caption{Detailed architecture of GRNet}
\begin{tabular}{ccc}
\toprule
Layer name & Input size & Output size \\
\hline
CBR   & $224\times224\times3$  & $224\times224\times64$   \\

BasicBlock & $224\times224\times64$  & $224\times224\times64$   \\

BasicBlock & $224\times224\times64$   & $224\times224\times64$   \\

BDB   & $224\times224\times64$  & $112\times112\times256$   \\

BasicBlock & $112\times112\times256$  & $112\times112\times256$   \\

BDB   & $112\times112\times256$  & $56\times56\times512$   \\

BasicBlock & $56\times56\times512$  & $56\times56\times512$  \\

AvgPool & $56\times56\times512$  & $1\times1\times512$  \\

Fully Connected Layer & $1\times1\times512$  & $1\times1\times4$  \\
\hline
\multicolumn{3}{c}{SoftMax}  \\
\bottomrule
\end{tabular}
\label{GRNet}
\end{table}

To further cut down the parameters and model size of GRNet, we use group convolution~\citep{howard2017mobilenets} to replace all $3\times3$ normal convolution. In general, we define a $D\times D\times C_{1}\times C_{2}$ convolution filter, where $D$ is the spatial dimension of the kernel assumed to be square. $C_{1}$ is the number of input channels, and $C_{2}$ is the number of output channels. The normal convolution can be defined as follows:
\begin{equation}
O_{m, n, c_{2}}=\sum_{i,j,c_{1}}D_{i, j, c_{1}, c_{2}}\cdot I_{m+i-1, n+j-1, c_{1}},
\label{eq:N-Conv}
\end{equation}
where $I$ and $O$ are input and output feature maps, respectively.

Group convolution splits the filters and feature maps into the same number of group in the channel direction. When the number of group is $g$, it is defined as:
\begin{equation}
\hat{O}_{m, n, c_{2}/g}=\sum_{i,j} \hat{D}_{i, j, g}\cdot I_{m+i-1, n+j-1, c_{1}/g} \;,
\label{eq:G-Conv}
\end{equation}
where $\hat{D}$ is the group convolution kernel of size $D\times D\times C_{1}/G \times C_{2}/G$, and $\hat{O}$ is the output feature map.

In this paper, we define the $g$ as the greatest common divisor of input and output channel numbers. If the dimensions of input and output are the same, the number of parameters in normal convolution is g times that of group convolution~\citep{howard2017mobilenets}.

\subsection{GYNet for Weld Information Recognition}
There are many foreground contents in weld image, including weld information, weld region and noises. The context of image is complicated because the scales of these contents vary greatly. To achieve accurate weld information recognition, a novel SCE module is proposed to enhance the contextual representation ability of extracted feature maps. In addition, we propose a narrow and efficient recognition network GYNet based on SCE module.

\begin{figure}[!t]
  \centering
  \includegraphics[width=5in]{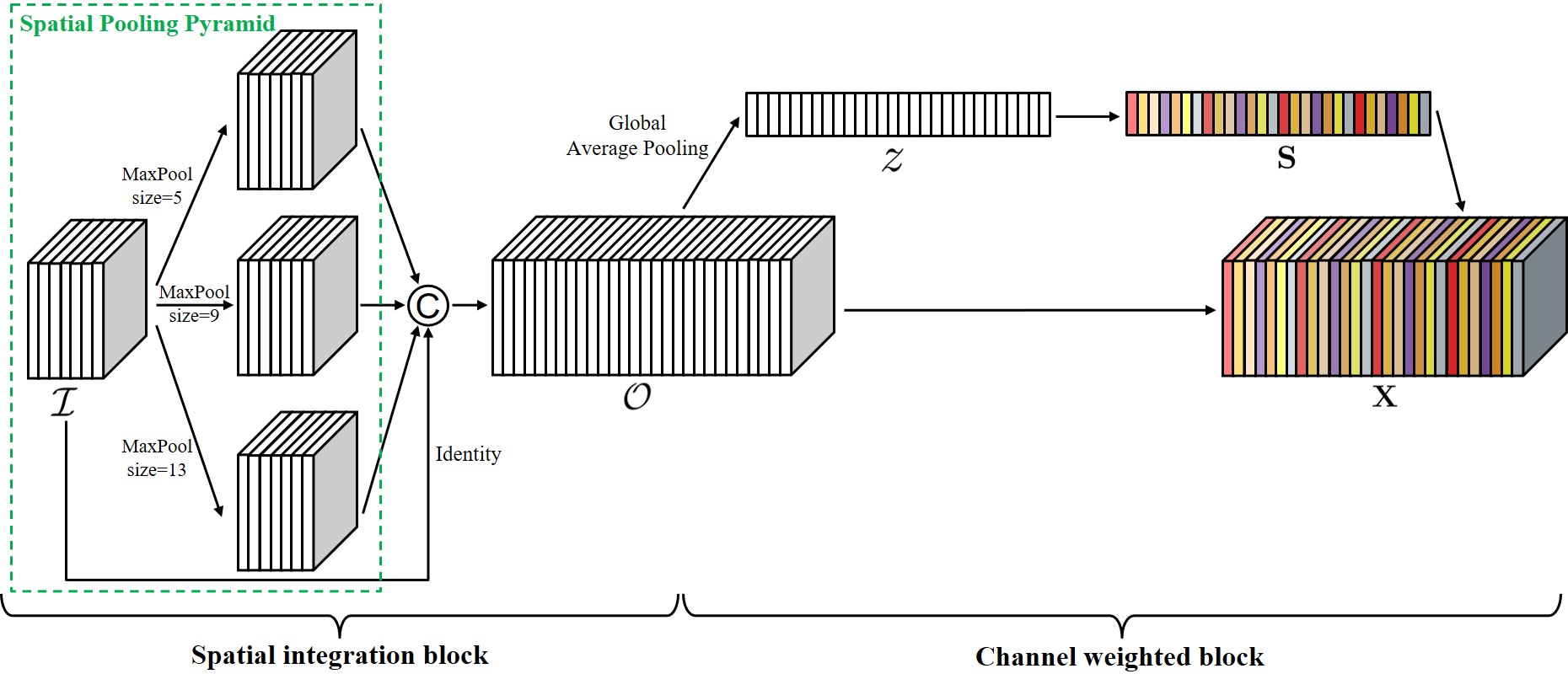}
  \caption{The internal structure of proposed SCE module. SCE is composed of spatial integration block and channel weighted block, and image is processed by them successively. The first block produces a spatial pooling pyramid and concatenates feature map among pyramid. The second block can obtain the importance of each channel and update feature maps based on weights.}
  \label{fig:SCE}
\end{figure}

\textbf{SCE Module.} To better utilize different scale features, we propose a novel SCE module and its detailed structure is shown in Fig.~\ref{fig:SCE}. SCE is composed of spatial integration block~\citep{bochkovskiy2020yolov4} and channel weighted block~\citep{hu2018squeeze}, and the feature maps are processed by these two blocks successively. Spatial integration block uses multiple pooling operations with different kernel sizes to obtain a spatial pooling pyramid, and it owns local and global features from different spatial scales. These features from diverse scales will be fused by concatenation at last, and the whole process can be formulated as follow:
\begin{equation}
\mathcal{O}=\mathcal{C}\left(\left.M_{i}(\mathcal{I})\right|_{i=1, 5, 9, 13}\right),
\label{eq:GAP}
\end{equation}
where $\mathcal{O}$ and $\mathcal{I}$ represent output and input feature map of spatial pooling pyramid, respectively. $\mathcal{C}(\cdot)$ is concatenation operation, and $M_{i}(\cdot)$ is maxpool with kernel size $i\times i$ ($i=1$ representing the identity branch). Spatial integration block fuses different receptive field information, which makes the obtained feature map capable of capturing diverse spatial features.

Channel weighted block is a type of attention mechanism method. It can learn the relationship between different channels and obtain the weight of each channel. Firstly, it uses global average pooling to generate channel descriptor $\mathcal{Z} \in \mathbb{R}^{1 \times 1 \times C}$ across spatial dimensions. $\mathcal{O} \in \mathbb{R}^{H \times W \times C}$ is input data, and the $c$-th element of $\mathcal{Z}$ is obtained as follow:
\begin{equation}
\mathcal{Z}_{c}=\frac{1}{H \times W} \sum_{i=1}^{H} \sum_{j=1}^{W} \mathcal{O}_{c}(i, j),
\label{eq:GAP}
\end{equation}
where $\mathcal{O}_{c}$ is the $c$-th feature map of $\mathcal{O}$. Then, the channel descriptor $\mathcal{Z}$ is excited to redefine the importance of each channel. Specifically, we employ a linear layer followed by ReLU layer and a linear layer followed by sigmoid layer, and the process can be described as:
\begin{equation}
\mathbf{S} = \sigma\left(\mathbf{W}_{2} \delta\left(\mathbf{W}_{1} \mathcal{Z}\right)\right),
\label{eq:excitement}
\end{equation}
where $\delta$ refers to the ReLU function, and $\sigma$ is Sigmoid function, $\mathbf{W}_{1} \in \mathbb{R}^{\frac{C}{r} \times C}$ and $\mathbf{W}_{2} \in \mathbb{R}^{C \times \frac{C}{r}}$. $r$ is a hyper-parameter that controls the model complexity, and it is set as 4 in this paper.

The final output of channel weighted block is calculated as:
\begin{equation}
\mathbf{X}=\mathfrak{\mathbf{F}}\left(\mathcal{O}, \mathbf{S}\right)=\mathcal{O} \cdot \mathbf{S},
\label{eq:channel-wise multiplication}
\end{equation}
where $\mathbf{X}\in \mathbb{R}^{H \times W \times C}$ and $\mathbf{F(\cdot)}$ refers to
channel-wise multiplication.

Channel weighted block is an adaptive adjuster whose function is to learn the importance of each channel information, and further can show which scale feature is more significant. Although multi-scale information is the basis of effective feature map, different scales make different contributions to the results. Especially when the sizes of recognized objects are similar, there is only one scale that is essential for final prediction theoretically. Compared with other foreground contents, the scale distribution of weld information is relatively consistent. Hence, channel weighted block is designed to weight different scale adaptively during network learning, and more significant channel, in other words, more meaningful scale feature would be assigned more weight.

Overall, the proposed SCE module improves the contextual representation ability of feature maps through integrating more information sources, and further weight them adaptively based on their importance. The effect of SCE will be discussed detailedly in Section 4.2.

\textbf{Architecture.} Inspired by Tiny-YOLO v3, we design a recognition network for weld information, and its detailed architecture is given in Table~\ref{tab:GYNet}. GYNet has the same numbers of Conv and MaxPool layer compared with Tiny-YOLO v3. More narrow model can decrease the parameters and FLOPs more directly. To obtain a smaller width backbone, we strictly limit the number of channels in each layer. Almost all layers are below 512 channels, and this design strategy makes network bring few burden on computation device. We embed SCE module at the tail of backbone to ensure it process more meaningful information, and make the enhanced features closer to the output layer for more accurate recognition results.


\begin{table}[!t]
\renewcommand{\arraystretch}{1.1}
\centering
\caption{Detailed architecture of GYNet}
\begin{tabular}{cccccc}
\toprule
Layer name & Type    & Filters &Size &Stride & Output size \\
\hline
S1         & Conv    & 16      &3    &1      & $416\times416\times16$   \\
\hline
S2         & MaxPool & --      &2    &2      & $208\times208\times16$   \\
\hline
S3         & Conv    & 32      &3    &1      & $208\times208\times32$   \\
\hline
S4         & MaxPool & --      &2    &2      & $104\times104\times32$   \\
\hline
S5         & Conv    & 64      &3    &1      & $104\times104\times64$   \\
\hline
S6         & MaxPool & --      &2    &2      & $52\times52\times64$   \\
\hline
S7         & Conv    & 64      &3    &1      & $52\times52\times64$   \\
\hline
S8         & MaxPool & --      &2    &2      & $26\times26\times64$   \\
\hline
S9         & Conv    & 128     &3    &1      & $26\times26\times128$   \\
\hline
S10        & MaxPool & --      &2    &2      & $13\times13\times128$   \\
\hline
S11        & Conv    & 256     &3    &1      & $13\times13\times256$   \\
\hline
S12        & MaxPool & --      &2    &1      & $13\times13\times256$   \\
\hline
S13        & Conv    & 512     &3    &1      & $13\times13\times512$   \\
\hline
S14        & SCE     & --      &--   &--     & $13\times13\times2048$   \\
\hline
S15        & Conv    & 128     &1    &1      & $13\times13\times128$   \\
\hline
S16        & Conv    & 256     &1    &1      & $13\times13\times256$   \\
\hline
S17        & Conv    & 135     &1    &1      & $13\times13\times135$   \\
\hline
S18        & Conv    & 128     &1    &1      & $13\times13\times128$   \\
\hline
S19        & UpSample& --      &--   &--     & $26\times26\times128$   \\
\hline
S20        & Concat  & --      &--   &--     & $26\times26\times256$   \\
\hline
S21        & Conv    & 256     &3    &1      & $13\times13\times256$   \\
\hline
S22        & Conv    & 135     &3    &1      & $13\times13\times135$   \\
\bottomrule
\end{tabular}
\label{tab:GYNet}
\end{table}

\section{Experiments}
To show the superiority of our framework for X-ray weld image signs recognition, experiment results and analysis are represented in this section. Firstly, the experimental setup including datasets, the implementation details and the evaluation metrics, is introduced in Section 4.1. Then, we validate the effectiveness of SCE module. Specifically, ablation studies are designed to show its necessity, and we visualize the weight values to prove aforementioned weight assignment mechanism. Finally, aiming at classification subtask and recognition subtask, we compare our proposed methods with the state-of-the-art models.

\subsection{Experimental Setup}
\textbf{Datasets.} We have obtained 897 digital X-ray weld images from the actual production workshops of special equipment companies. All images have been annotated carefully by professionals. We build two datasets for training/testing classification and recognition.

For the classification subtask, to make the cross mark more eye-catching in the image, we divide the image into $416\times416$ pixels multiple sub-images as input. We use the flip and minor operation to augment our dataset for obtaining a more robust network. At the end, we have 3588 images in cross mark classification dataset, and it is randomly divided into training set, validation set and testing set according to the ratio of 8:1:1. Cross mark classification dataset has four classes to represent the direction of images.

For the recognition subtask, we resize the whole image to $416\times416$ pixels for adapting to the normal input size of YOLO. The number of weld information classes is 40, which is relatively large, and the condition of whole image is very complex. So, we use more complicated augmentation methods by combining changing brightness and contrast, using Gaussian blur and rotating way. Each original image is augmented two or three times randomly, and finally obtain 3550 images, which are randomly divided into 3145 images in training set and 355 images in test set.

\textbf{Implementation Details.} We conduct our all experiments on a i7-8700K CPU and a single NVIDIA GeForce GTX1070Ti GPU. All models are based on deep learning framework PyTorch. In cross sign classification experiments, we choose stochastic gradient descent optimizer with 0.9 momentum parameter and 0.0005 weight decay. The initial learning rate and total epochs are set as 0.1 and 80, respectively. The step policy is used to divide initial learning rate in 10 by each 50 epochs. Label smoothing strategy is introduced to optimize the classification training process. In information recognition experiments, the series of YOLO networks are trained by SGD optimizer with 0.9 momentum parameter and 0.0005 weight decay as well. The initial learning rate and total epochs are set as 0.001 and 300, respectively. Learning scheduler utilizes LambdaLR strategy. The state-of-the-art methods are trained and tested based on MMdetection~\citep{chen2019mmdetection} which is an open source 2D detection toolbox, and all related hyper-parameters are adjusted to the optimal.

\textbf{Evaluation Metrics.} We adopt mean average precision (mAP), Recall, floating-point of operations (FLOPs), parameters (Params) and frame per second (FPS) as evaluation metrics to evaluate the proposed network comprehensively. mAP and Recall are used to show the detection performance, while the rest of metrics are used to represent the computation complexity and speed property. Relevant metrics are defined as:

\begin{equation}
Precision=\frac{TP}{TP + FP},
\end{equation}
\begin{equation}
Recall=\frac{TP}{TP + FN},
\end{equation}
and the involved concepts TP, FP, FN ~\citep{2020A} are explained as follow.

\begin{itemize}
\item True positive (TP): The number of objects that are detected correctly.
\item False positive (FP): The number of incorrect detection including nonexistent and misplaced predictions.
\item False negative (FN): The number of objects that are undetected successfully.
\end{itemize}

mAP is a metric used to evaluate comprehensive ability over all classes. It is simply the average AP over all classes, and AP can be computed as the area of Precision $\times$ Recall curve with axles.

Moreover, to compare the computation complexity of different networks, time complexity, i.e., FLOPs and space complexity, i.e., Params are chosen to show the difference between different methods. In addition, we use FPS to show the speed during inference stage, and the results of FPS are the average of 350 testing images in this paper.

\subsection{Effectiveness of the SCE module}

\textbf{Ablation Studies.} To explore the importance of each module in our GYNet, we design a series of ablation studies, and the obtained results are shown in Table~\ref{tab:ablation}. Based on similar backbone, all combinations are much the same in terms of recognition speed and computations. Although the introduction of both two blocks brings a slight increase in the number of parameters and model size, the recognition ability of our method has a great improvement, attaining 90.0 mAP and 88.8$\%$ recall. However, the result improvement is extremely limited when either of them is used alone. We attribute the superior performance of our SCE module to its ability of weighting the information after feature fusion.

\begin{table*}[!t]
	\renewcommand{\arraystretch}{1.2}
	\renewcommand\tabcolsep{2.5pt}
	\caption{The comparison results from different combinations of spatial integration block (SIB), channel weighted block (CWB).}
	\centering
	\label{tab:ablation}
	\begin{tabular}{cc|c|ccccccc}
		\toprule
	    & SIB                         & CWB                       & mAP            & Recall($\%$)  & FPS      & FLOPs(G)      & Params(M)     & Size(MB)     \\
		\midrule
		&--                           &--                         & 86.4           & 87.4          & \textbf{195.0}    & \textbf{2.8}  & \textbf{2.6}  & \textbf{10.7} \\
		&$\surd$                      &--                         & 87.2           & 87.5          & 165.8    & \textbf{2.8}           & 2.8           & 11.5  \\
		&--                           &$\surd$                    & 87.0           & 87.5          & 191.5    & \textbf{2.8 }          & 2.8           & 11.2  \\
		&$\surd$                      &$\surd$                    & \textbf{90.0}  & \textbf{88.8} & 176.1    & \textbf{2.8}           & 4.9           & 19.9  \\
		\bottomrule
	\end{tabular}
\end{table*}

\begin{figure}[!t]
	\centering
	\includegraphics[width=4.2in]{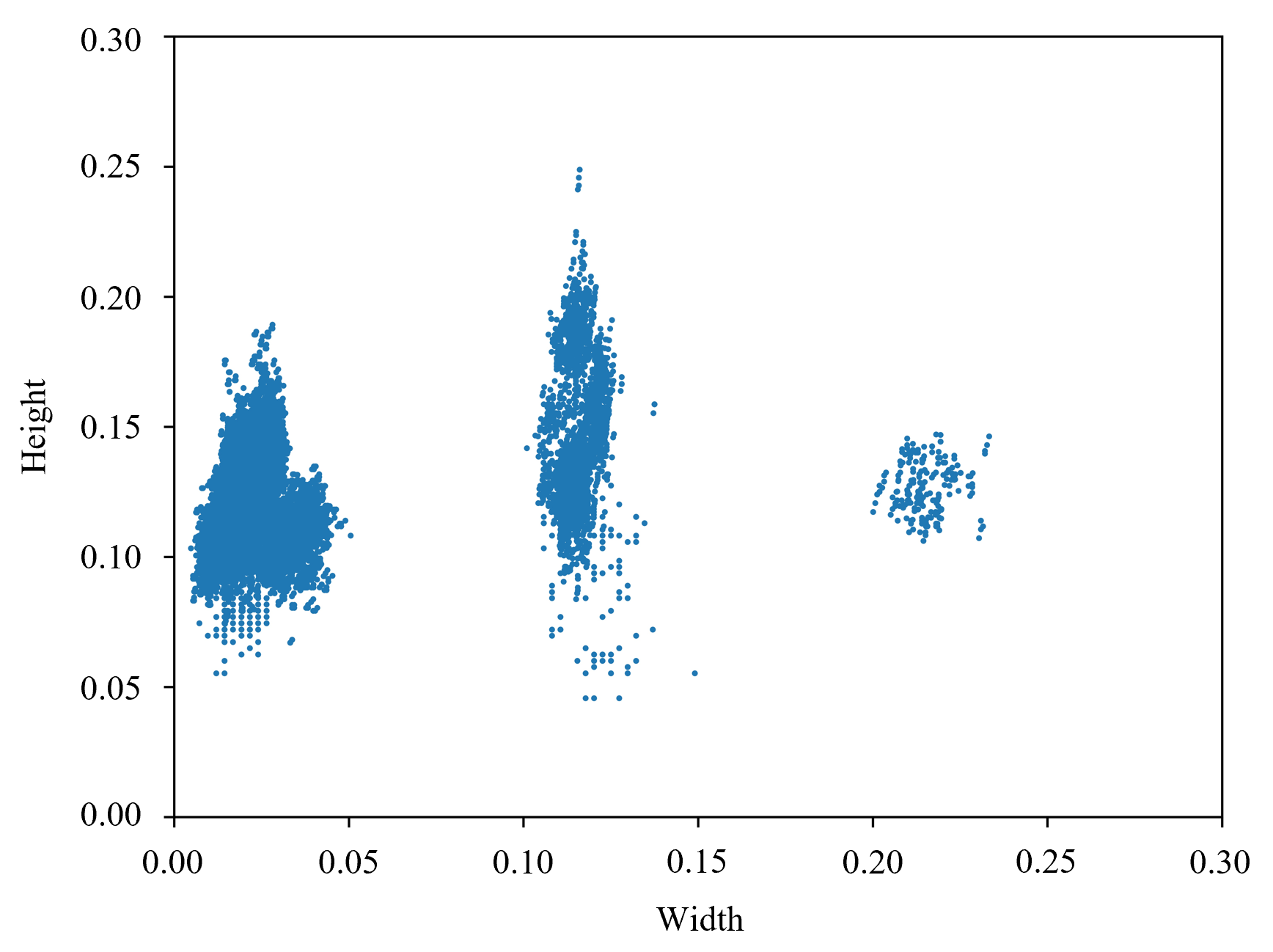}
	\caption{The scale distribution of weld information.}
	\label{fig:label distribution}
\end{figure}

\begin{figure}[!t]
	\centering
	\includegraphics[width=4.2in]{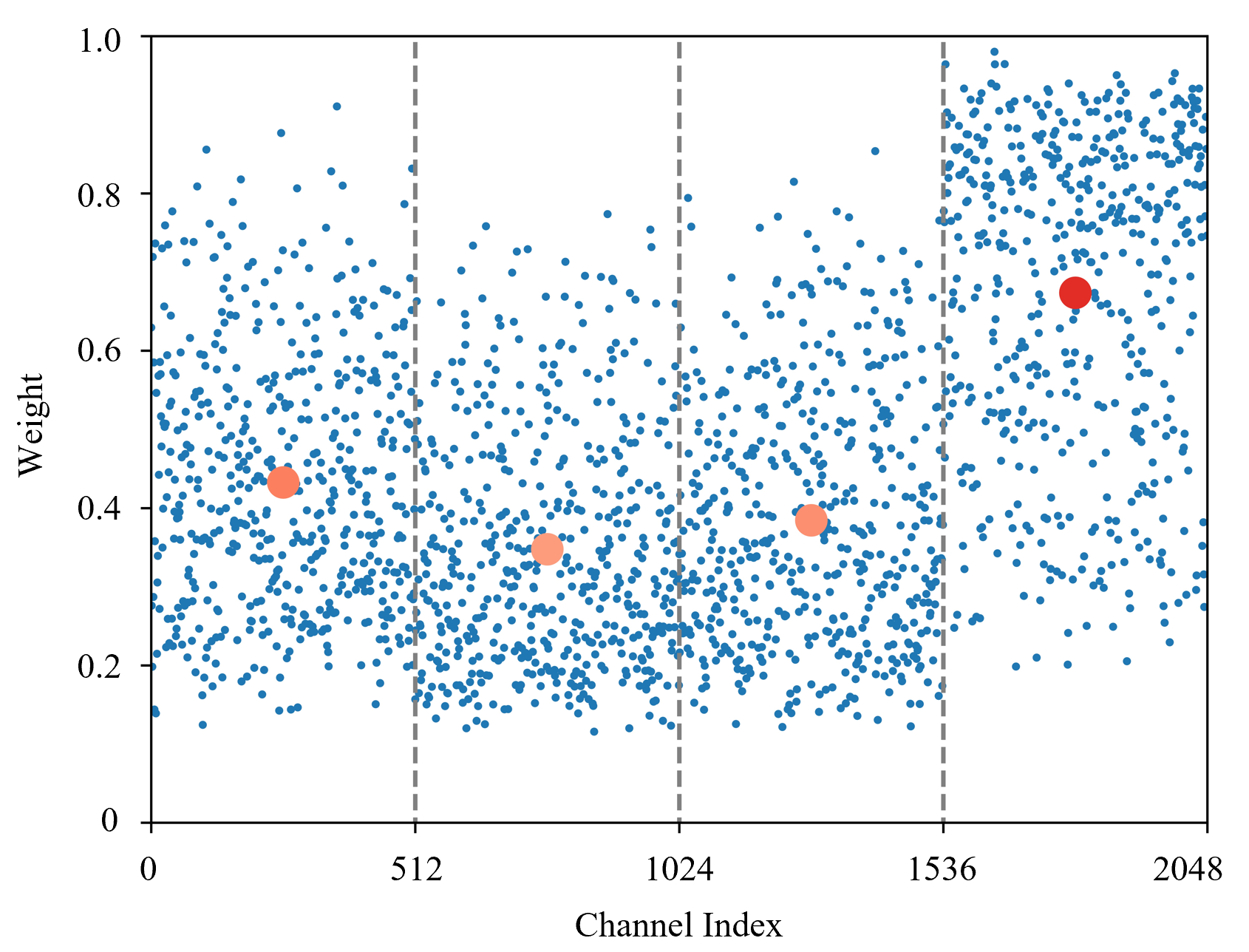}
	\caption{The weight values produced by channel weighted block of SCE module. The blue dots represent the weight values of each channel, and the red dots show the average weight values of each channel interval.}
	\label{fig:SCE weight}
\end{figure}

\textbf{Validation of Weight Assignment Mechanism.}  To intuitively observe the scale distribution of weld information, we normalize the width and height of weld information relative to weld image size, and its scale distribution is shown in Fig.~\ref{fig:label distribution}. The scale of weld information is consistent, while widths and heights are almost all less than 0.25, and it means that weld information belongs to a relatively small scale space. To validate the reasonability of weight assignment mechanism, we visualize the weights produced by channel weighted block as shown in Fig.~\ref{fig:SCE weight}. We divide channels into four parts, corresponding to four features scales of the spatial pooling pyramid. The blue dots indicate the weights assigned to each channel by channel weighted block. Red dots indicate the average weight of each channel interval, and color deepness reflects the size of the average value. It can be observed that SCE module assigns almost average weight of 0.4 to the channels of first three intervals, and about average weight of 0.7 to the last interval which comes from the identity branch. Maxpool operation with large kernel size would weaken local feature, and it is not favorable for the recognition of small scale object. Hence, SCE module assigns more weights to identity branch, while treating other scale sources as less important contributions. This results show that the proposed SCE module can adaptively weight each feature source based on multi-scale feature fusion.

\subsection{Comparisons with State-of-the-art Models}
\begin{figure}[!t]
	\centering
	\includegraphics[width=4.2in]{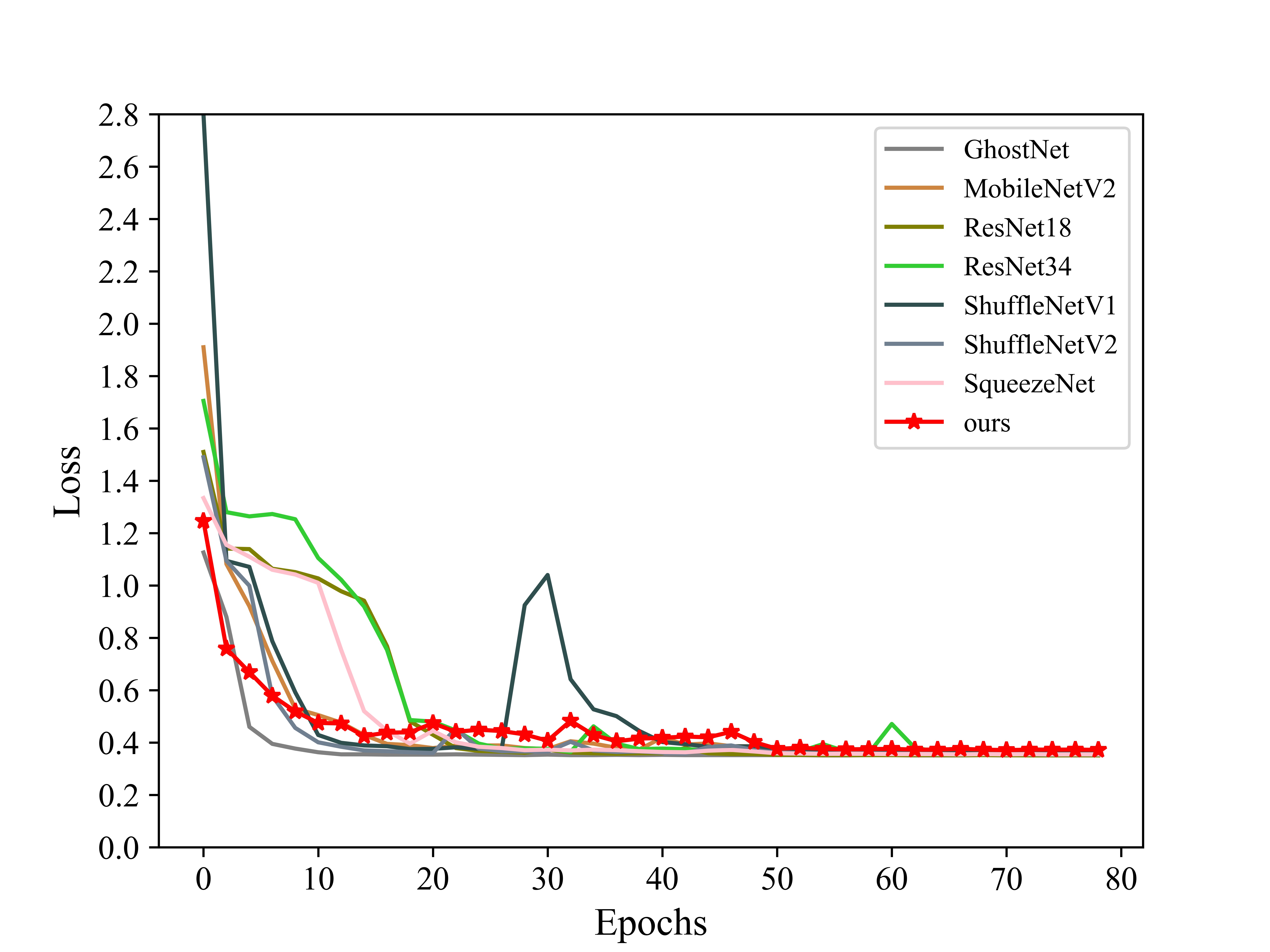}
	\caption{The training loss curve comparison of different classification methods.}
	\label{fig:loss}
\end{figure}

\textbf{The Classification of Cross Mark.} In order to validate the performance of our GRNet in cross mark classification dataset, we introduce many advanced classification networks, including classical models like ResNet-18~\citep{he2016deep} and ResNet-34~\citep{he2016deep}, and lightweight classification networks such as ShuffleNetV1~\citep{zhang2018shufflenet}, ShuffleNetV2~\citep{2018ShuffleNet}, MobileNetV2~\citep{2018MobileNetV2}, SqueezeNet~\citep{iandola2016squeezenet}, GhostNet~\citep{han2020ghostnet}. The loss curves of above networks during training process are presented in Fig.~\ref{fig:loss}, and we can observe that the performance of all models tends to be stable after 50 epochs. It is worth noting that the loss values of our GRNet have less fluctuation at the beginning, and come down to very low and similar loss values eventually compared with other methods.

\begin{table}[!t]
	\renewcommand{\arraystretch}{1.2}
	\renewcommand\tabcolsep{0.5pt}
	\caption{The classification results in comparison with state-of-the-art methods.}
	\centering
	\label{tab:GRNet}
	\begin{tabular}{l|ccccc}
		\toprule
		Models                                    & Accuracy($\%$)   & FLOPs(G)      & Madds(G)     & Params(M)    & Size (MB)    \\
		\midrule
		ShuffleNetV1~\citep{zhang2018shufflenet}   & 99.7             & 2.2           & 4.3          & 0.9          & 3.9 \\
		ShuffleNetV2~\citep{2018ShuffleNet}        & 99.4             & 2.2           & 4.4          & 1.3          & 5.2 \\
		MobileNetV2~\citep{2018MobileNetV2}        & 99.7             & 2.4           & 4.8          & 2.3          & 9.2 \\
		SqueezeNet~\citep{iandola2016squeezenet}   & 99.7             & 2.7           & 5.2          & 0.7          & 3.0 \\
		GhostNet~\citep{han2020ghostnet}           & 96.1             & \textbf{0.2}  & \textbf{0.3} & 3.9          & 15.8 \\
		ResNet-34~\citep{he2016deep}               & \textbf{100.0}   & 56.9          & 113.8        & 21.3         & 85.2 \\
		ResNet-18~\citep{he2016deep}               & 99.7             & 27.3          & 54.5         & 11.2         & 44.7 \\
\hline
		GRNet(ours)                                  & 99.7             & 1.1           & 2.1          & \textbf{0.2} & \textbf{0.8} \\
		\bottomrule
	\end{tabular}
\end{table}

From Table~\ref{tab:GRNet}, it can demonstrate that our GRNet outperforms all other methods, while it achieves 99.7$\%$ accuracy with the least 0.2M Params and 0.8 MB model size. Although the accuracy of ResNet-34 is 100\%, its parameters are 112 times that of our method, and its FLOPs and Madds are the most, 52 and 53 times that of our method, respectively. ShuffleNetV1, MobileNetV2, SqueezeNet and ResNet-18 also obtain the same accuracy compared with our method, but they are not lightweight enough. Specifically, ResNet-18 has $25\times$ FLOPs and $59\times$ Params than our GRNet. In comparison with the other three lightweight models, our GRNet also achieves greater lightweight performance, while it only has about 50$\%$ FLOPs, and 8$\%$$\sim$26$\%$ Params. Therefore, we can conclude that our method can classify cross mark accurately with lower computational complexity and fewer resources. GhostNet has lowest FLOPs and multiply-add operations (Madds) because of its particular design of reusing feature maps. But its accuracy is only 96.1$\%$ and the number of parameters is $20.6\times$ than our method.

\begin{table}[!t]
	\renewcommand{\arraystretch}{1.2}
	\renewcommand\tabcolsep{0.5pt}
	\caption{The recognition results in comparison with state-of-the-art methods. $\dagger$ indicates that network combines with SCE module.}
	\centering
	\label{tab:GYNet-Comparison}
	\begin{tabular}{l|cccccc}
		\toprule
		Methods                     	                 & mAP           & FPS            & FLOPs(G)    & Madds(G)  & Params(M)    & Size(MB)\\
		\midrule
		Tiny-YOLO v3~\citep{redmon2018yolov3}             & 87.3          & 143.3          & 5.5        & 11.0       & 8.8          & 35.1 \\
		YOLO v3~\citep{redmon2018yolov3}                  & 90.4          & 25.6           & 65.8       & 131.6      & 61.7         & 247.3 \\
		RetinaNet~\citep{lin2017focal}                    & 87.3          & 20.1           & 37.5       & 75.0       & 36.9         & 148.8 \\
		Faster R-CNN~\citep{2017Faster}                   & 90.2          & 18.1           & 46.7       & 93.4       & 41.3         & 166.5 \\
		Cascade R-CNN~\citep{cai2018cascade}              & 90.1          & 12.2           & 74.4       & 148.8      & 69.1         & 277.4 \\
		Libra R-CNN~\citep{pang2019libra}                 & 90.0          & 17.2           & 46.9       & 93.8       & 41.6         & 167.5 \\
		Dynamic R-CNN~\citep{zhang2020dynamic}            & 89.5          & 18.2           & 46.9       & 93.8       & 41.5         & 166.5 \\
		Tiny-YOLO v3$\dagger$                             & 87.9          & 114.5          & 5.8        & 11.6       & 17.9         & 71.8 \\
		YOLO v3$\dagger$                                  & \textbf{91.0} & 25.0           & 66.4       & 132.8      & 71.7         & 287.2 \\
\hline
		GYNet(ours)                                       & 90.0          & \textbf{176.1} & \textbf{2.8} & \textbf{5.6} & \textbf{4.9} & \textbf{19.9} \\
		\bottomrule
	\end{tabular}
\end{table}

\begin{figure*}[htbp!]
	\centering
	\subfigure{
		\includegraphics[width=4.7in]{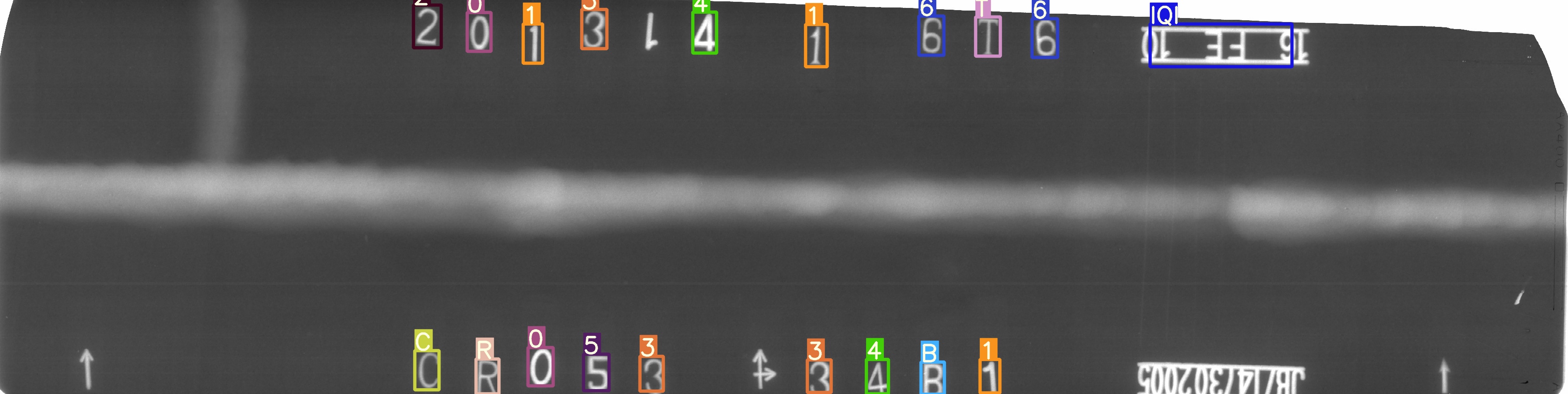}
	}
	\subfigure{
		\includegraphics[width=4.7in]{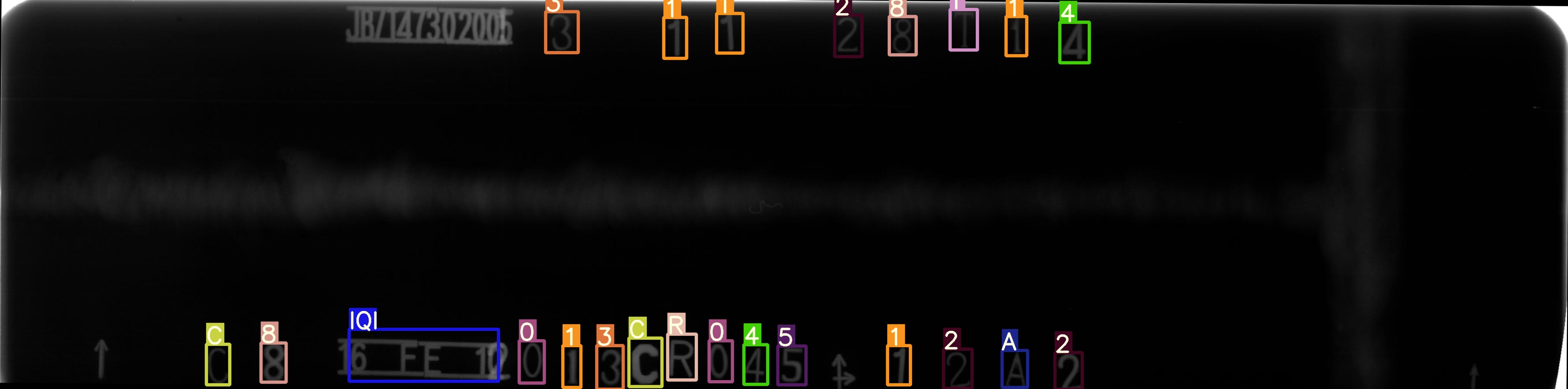}
	}
	\subfigure{
		\includegraphics[width=2.3in]{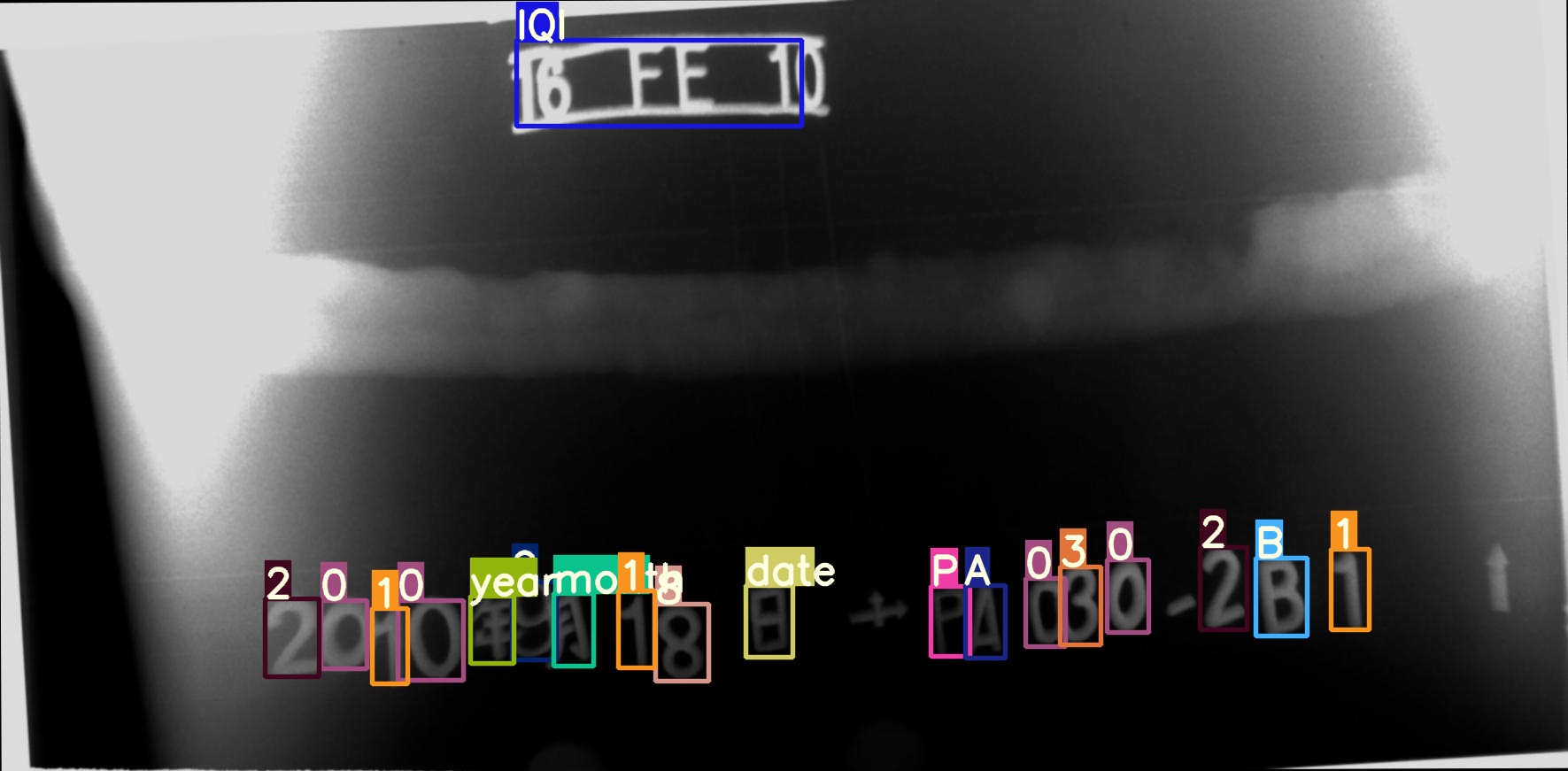}
	}
	\subfigure{
		\includegraphics[width=2.3in]{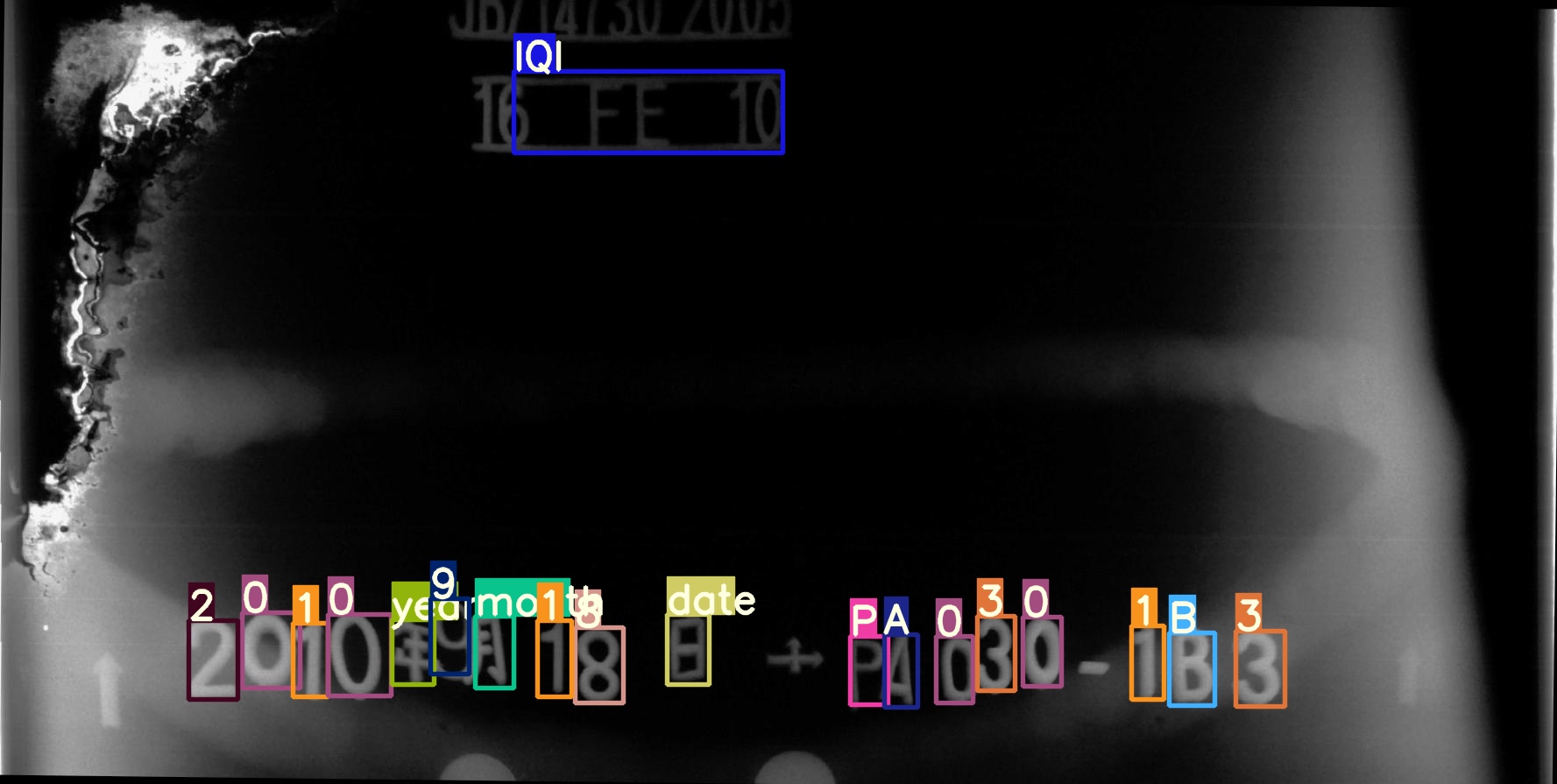}
	}
	\subfigure{
		\includegraphics[width=2.3in]{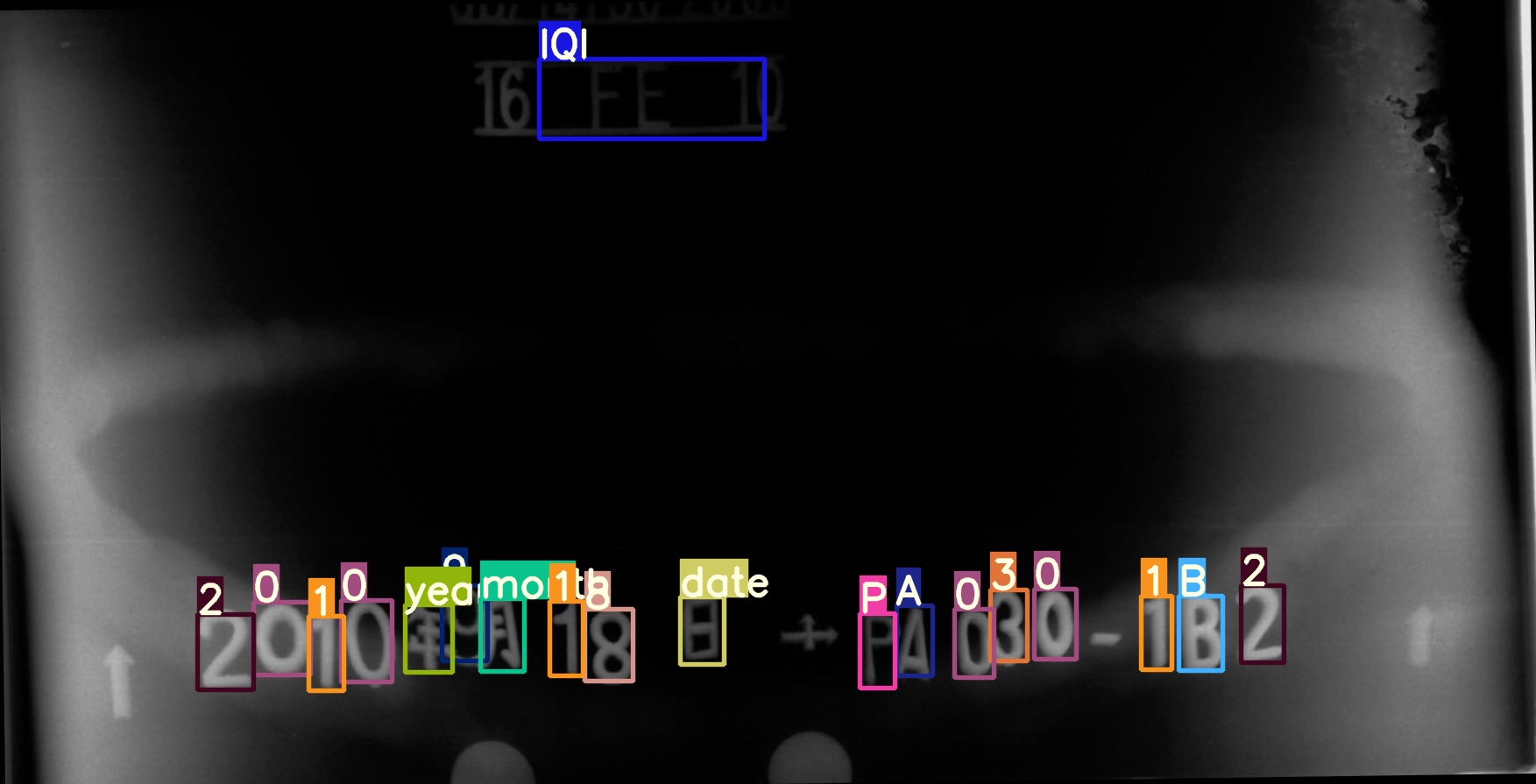}
	}
	\subfigure{
		\includegraphics[width=2.3in]{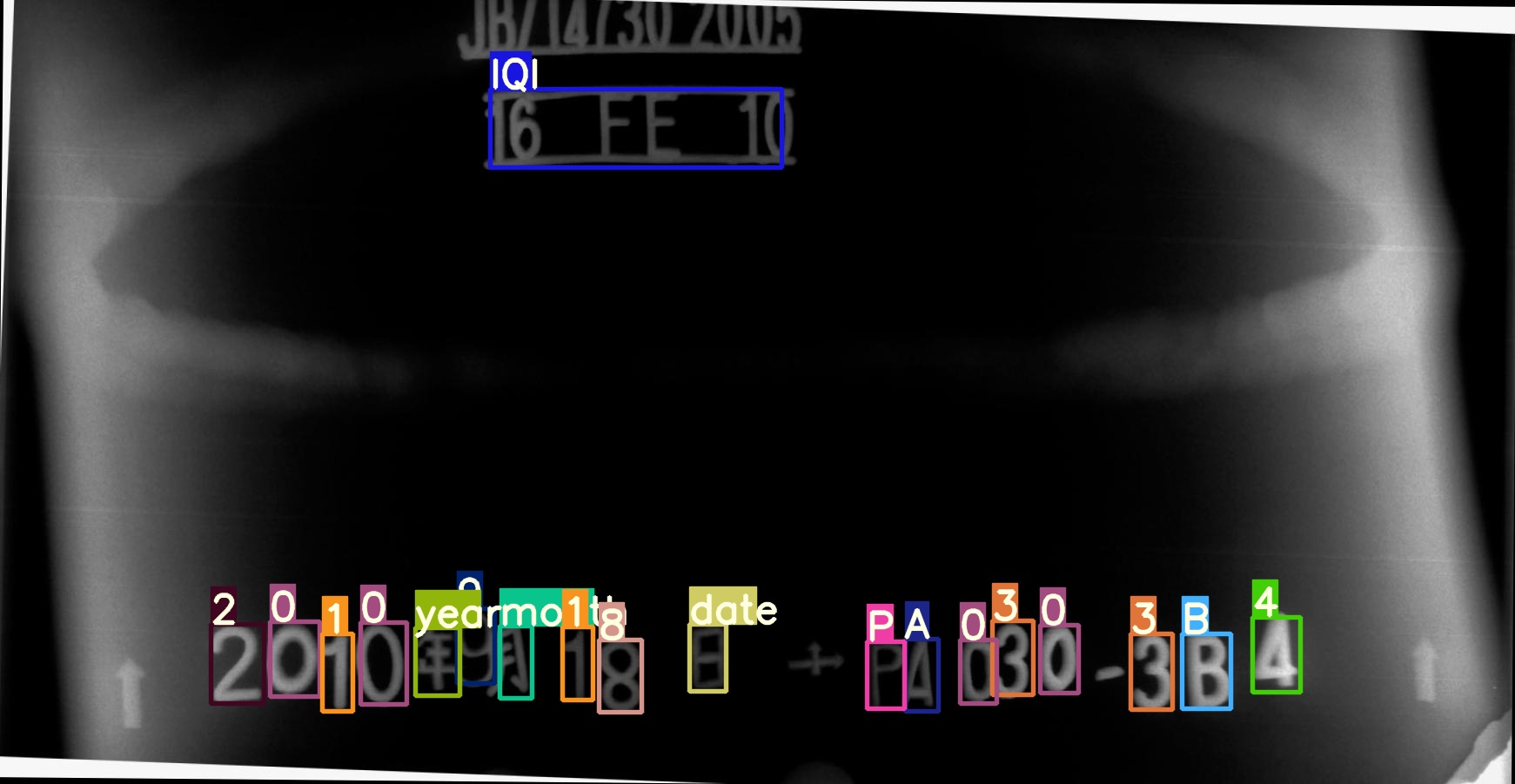}
	}
	\caption{Visualization results of weld information recognition produced by GYNet. GYnet can accurately recognize weld information for different types of X-ray images, even when the brightness of image is extremely low. }
	\label{fig:output results}	
\end{figure*}

\textbf{The Recognition of Weld Information.} Based on the MMdetection, we compare our method with many state-of-the-art models using ResNet-50 as backbone, such as RetinaNet~\citep{lin2017focal}, Faster R-CNN~\citep{2017Faster}, Cascade R-CNN~\citep{cai2018cascade}, Libra R-CNN~\citep{pang2019libra}, Dynamic R-CNN~\citep{zhang2020dynamic}, and all related parameters have been set to make all models perform best. The comparison results are shown in Table~\ref{tab:GYNet-Comparison}. The Params of our GYNet is only 4.9M, which is 55.7$\%$ of Tiny-YOLO v3 and 7.9$\%$ of YOLO v3. Such small number of parameters makes the model size only 19.9 MB, 15.2 MB smaller than the famous compressed network Tiny-YOLO v3. In addition, GYNet has the fastest speed with 176.1 FPS, which is $1.2\times$ faster than Tiny-YOLO v3 and $6.9\times$ faster than YOLO v3. Under such deep lightweight optimization, our method still achieves 90.0 mAP, and is much higher than its baseline Tiny-YOLO v3 and famous one-stage network RetinaNet, while only 0.4 points lower than YOLO v3. Classical two-stage CNN models Faster R-CNN, Cascade R-CNN, and Libra R-CNN have similar performance compared with our GYNet, but their recognition speeds are all below 30 FPS, far from meeting the actual requirements. Furthermore, their Params and model size are overweight for normal hardware. We combine SCE module with YOLO v3 and Tiny-YOLO v3, and their performance has been improved, which validates the effectivenss of SCE module as well. The visualization recognition results of GYNet are shown in Fig.~\ref{fig:output results}, and we can observe that our GYNet is capable of dealing with weld images with various types and lighting conditions.

\section{Conclusion}
In this paper, we propose a high-performing lightweight framework for signs recognition of weld images, in which GRNet and GYNet are connected to complete whole task. Aiming at classification of cross mark, GRNet is presented with a well-designed backbone to compress model. For weld information, a new architecture with a novel SCE module is designed for GYNet, in which the SCE module integrates multi-scale features, and assigns adaptively weights to different scale sources. Experiments show that our signs recognition framework obtains high prediction accuracy with tiny parameters and computations. Specifically, GRNet achieves 99.7\% accuracy with only 0.8 MB model size and 1.1 GFLOPs, which is 1.8\% and 4\% of ResNet-18, respectively. GYNet achieves 90.0 mAP on recognition dataset, 2.7 points higher than Tiny-YOLO v3, and its FPS is 176.1, $1.2\times/6.9\times$ faster than Tiny-YOLO v3/ YOLO v3. In the future, we will focus on the further optimization of algorithm, and the application on embedded platforms (Raspberry Pi and Jetson Nano) to reduce hardware costs.


\bibliographystyle{elsarticle-harv}
\bibliography{mybibfile}
\end{document}